**Facilitating connected autonomous vehicle operations using space-weighted information fusion and deep reinforcement learning based control**

Jiqian Dong[a], Sikai Chen[b*], Yujie Li[a], Runjia Du[a], Aaron Steinfeld[c], Samuel Labi[a]

a. Center for Connected and Automated Transportation (CCAT), and Lyles School of Civil Engineering, Purdue University, W. Lafayette, IN 47907.
b*. Center for Connected and Automated Transportation (CCAT), Lyles School of Civil Engineering, Purdue University, W. Lafayette, IN 47906, and Robotics Institute, School of Computer Science, Carnegie Mellon University, Pittsburgh, PA (corresponding author, phone: 213-806-0141; e-mail: chen1670@purdue.edu; sikaichen@cmu.edu
c. Robotics Institute, School of Computer Science, Carnegie Mellon University, Pittsburgh, PA 15213.

**Abstract**
The connectivity aspect of connected autonomous vehicles (CAV) is beneficial because it facilitates dissemination of traffic-related information to vehicles through Vehicle-to-External (V2X) communication. Onboard sensing equipment including LiDAR and camera can reasonably characterize the traffic environment in the immediate locality of the CAV. However, their performance is limited by their sensor range (SR). On the other hand, longer-range information is helpful for characterizing imminent conditions downstream. By contemporaneously coalescing the short- and long-range information, the CAV can construct comprehensively its surrounding environment and thereby facilitate informed, safe, and effective movement planning in the short-term (local decisions including lane change) and long-term (route choice). The current literature contains useful information on CAV control approaches that use only local information sensed from the proximate traffic environment but relatively little guidance on how to fuse this information with that obtained from downstream sources and from different time stamps, and how to use the fused information to enhance CAV movements. In this paper, we describe a Deep Reinforcement Learning based approach that integrates the data collected through sensing and connectivity capabilities from other vehicles located in the proximity of the CAV and from those located further downstream, and we use the fused data to guide lane changing, a specific context of CAV operations. In addition, recognizing the importance of the connectivity range (CR) to the performance of not only the algorithm but also of the vehicle in the actual driving environment, the study carried out a case study. The case study demonstrates the application of the proposed algorithm and duly identifies the appropriate CR for each level of prevailing traffic density. It is expected that implementation of the algorithm in CAVs can enhance the safety and mobility associated with CAV driving operations. From a general perspective, its implementation can provide guidance to connectivity equipment manufacturers and CAV operators, regarding the default CR settings for CAVs or the recommended CR setting in a given traffic environment.

**Introduction**
Motivated by the challenges associated with safety and mobility in the traditional highway environment, and spurred by ongoing advancements and opportunities in information and communications technologies, government agencies and the research community continue to seek guidance on a number of aspects associated with vehicle connectivity and automation (AASHTO, 2018; FHWA, 2018; USDOT, 2019). These aspects include the enabling technologies, demand assessment, impact assessment and policy issues (including legal and ethics), human factors, infrastructure readiness, operations and controls, and implementation of this new generation of vehicles (USDOT, 2019). The quest for guidance has proceeded with due recognition (or at least, with the expectation) that in any given future era, the scope and profoundness of each of these aspects will be heavily influenced by the prevailing level of market penetration and level of autonomy. As is the case with any new transportation stimulus including innovations, it is imperative to assess the resultant effects using a carefully-designed portfolio of performance outcomes (FHWA, 2019; Sinha and Labi, 2007; World Bank, 2005). In the context of automated and connected vehicle operations, such outcomes may be viewed from the perspective of the impact type (safety, mobility, privacy, equity, for example), impact direction (costs and benefits), and the affected stakeholder (the transportation agency, road user, and the community) (Lioris et al., 2017; Litman, 2014; TRB, 2019, 2018).

The optimism associated with the prospective benefits of automated and connected vehicle operations is somewhat tempered by the realization that it will take a long time for fully automated vehicles to dominate the traffic stream (Litman, 2014). As such, the mostly likely situation will be the existence of a so-called transition period where the road will be shared by CAVs and HDVs, a situation often referred to as mixed or heterogeneous traffic (Chen et al., 2017, 2016; Li et al., 2020a; 2020b). Li et al. (2020) and Zhou and Zhu (2020) used theoretical analysis and a case study to demonstrate that a higher share of CAV in the traffic stream can cause significant shifts in roadway capacity and traffic flow patterns, respectively. It is essential to carefully model and study the effect of mixed traffic on the transportation system before CAVs are widely implemented and adopted; that way, it will be easier to reap the benefits of CAVs. In order to estimate the full range of CAV operational impacts in terms of the outcomes, some knowledge of the underlying CAV technology, including its capabilities, is useful. A common thread in CAV technology is (a) the elimination of the human from the driving task (a feature of vehicle autonomy), and (b) the ability to communicate with other road users and the infrastructure (a feature of vehicle connectivity). Each of these two innovations, by themselves, are expected to result in disruptive and far-reaching consequences on the traditional highway transportation operating environment. With regard to vehicle automation, the National Highway Traffic Safety Administration (NHTSA) has defined the categories of vehicle automation (NHTSA, 2016). It is anticipated that the entry of fully autonomous vehicles (Level 4 and Level 5 automation) in the market will cause drastic disruptions on the landscape of highway transportation (USDOT, 2019). Connectivity is often discussed as a part of the "Internet of Things" concept (Ashton, 2009; Guerrero-Ibanez et al., 2015; Ha et al., 2020a),wherein elements of a system share information that are useful for making decisions that enhance system efficiency, capacity, and safety. In the transportation domain, connectivity includes vehicle-to-vehicle (V2V), vehicle-to-infrastructure (V2I), vehicle-to-cloud (V2C), and other forms of vehicle-to-external (V2X) communication capabilities. It has been postulated that connectivity technology will greatly benefit the safety and efficiency of vehicle operations (Elliott et al., 2019) by promoting greater awareness of the driving environment, and therefore, will facilitate proactive actions to enhance driving performance (FHWA, 2015).

We present Figure 1 (below) to illustrate this concept. In the figure, the vehicle of interest (also termed the "ego" vehicle) is denoted in red color. The figure presents a situation where the ego vehicle is presented an opportunity to exploit its connectivity capabilities to make safer rational driving decisions. If the ego vehicle is capable of accessing information only from other vehicles within its immediate vicinity (that is, its sensing range), it will likely decide to stay in its lane (Lane 1) because the (white) vehicle ahead of the ego vehicle is moving at a higher speed compared to the vehicle in lane 2 (blue vehicle), all in the vicinity of the ego vehicle. Assume that further downstream in Lane 1, there exists an imminent hazard associated with different infrastructure settings or traffic situation (for example, a crash site, entry ramp, disabled vehicle, or workzone). If the ego vehicle's sources of information are limited to its local area only, it will be unable to characterize these imminent conditions downstream, and will continue driving until it reaches the threat, whereupon it will need to decelerate sharply or undertake some evasive maneuver. On the other hand, if the ego vehicle's sources of information include connected vehicles sources located further downstream, then it will be able to sense the imminent situation well before it reaches the threat location, and therefore will make an early decision to decelerate while in Lane 1, merge into lane 2, or both.

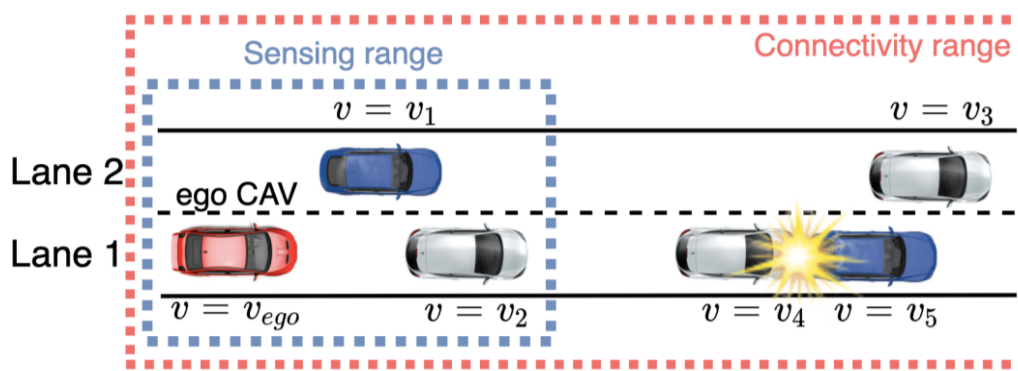

Figure 1 Conceptual ranges of sensing and connectivity, in lane-changing situation (the "Ego" vehicle or the CAV of interest, is colored red)

It has been postulated that the combined effect of automation and connectivity will yield benefits that exceed the sum of the individual benefits of these technologies. In this paper, we do not investigate this hypothesis or measure the synergistic effect of these two technologies. Nevertheless, we duly recognize that the coupling of connectivity and automation can accentuate vastly the benefits of the latter. Due to recent and ongoing advancements in control theory and Artificial Intelligence (AI) techniques, not only have automation and connectivity of highway vehicles become more sophisticated individually but also their integration has become increasingly feasible. With regard to CAV control technologies, there currently exist two main directions – optimization-based control and intelligent control. Optimization-based control seeks to generalize the physical driving task as a minimization (or maximization) objective function with multiple constraints, and then solve for the control input values. Several recent research efforts have successfully solved problems including CAV's trajectory planning (Yu et al., 2019), multi-platoon cooperative control (Li et al., 2019; Du et al., 2020a), joint control of CAV and traffic signals (Feng et al., 2018). However, as driving environment becomes increasingly complicated (for example, as the number of vehicles increases and as optimization-based control methods become highly non-convex), it has become difficult to reach solutions in linear time. This is inimical to CAV operations where the capability to make instant decisions is imperative.

On the other hand, the intelligent controller concept consists of Deep Learning (DL), Reinforcement Learning (RL), and other AI techniques that leverage the universal functional approximation ability of Deep Neural Network (DNN) models to approximate complicated decision processes. The advantage is intuitive: when the model is well trained, the inference time is fixed and short. The application of DL has led to significant improvements in civil engineering and transportation domains and has seen several applications in traffic flow prediction (Cui et al., 2019; Hou and Edara, 2018; Huang et al., 2019; Polson and Sokolov, 2017), infrastructure management (Attoh-Okine, 1999; Roberts and Attoh-Okine, 1998), smart grids (Fainti et al., 2017), and autonomous vehicle operations, particularly, in characterizing the driving environment (Zhu et al., 2017), route planning, and operational decision making (Schwarting et al., 2018; Veres et al., 2011; Ha et al., 2020b), and vehicle-to-vehicle communications (Ye and Li, 2018). The other sibling technique, Reinforcement Learning, is typically used to model Markov Decision Processes (MDP) and enable agents to identify optimal policies for interacting with a dynamic environment (Kaelbling et al., 1996). The integration of DL and RL, referred to as Deep Reinforcement Learning (DRL), greatly enhances the representation power in RL models and facilitates the analysis of extremely complicated scenarios (Mousavi et al., 2018; Dong et al., 2020).

In discussing DRL-based control of CAVs, it is essential to recognize the various levels of CAV decision making. This is because the level of a specific CAV operations task will influence the design of the controller intended to address that task. CAV driving operations can be placed into three classes based on the level of the driving decision (LODD) (Chen, 2019; S. Chen et al., 2020), as shown in Figure 2.

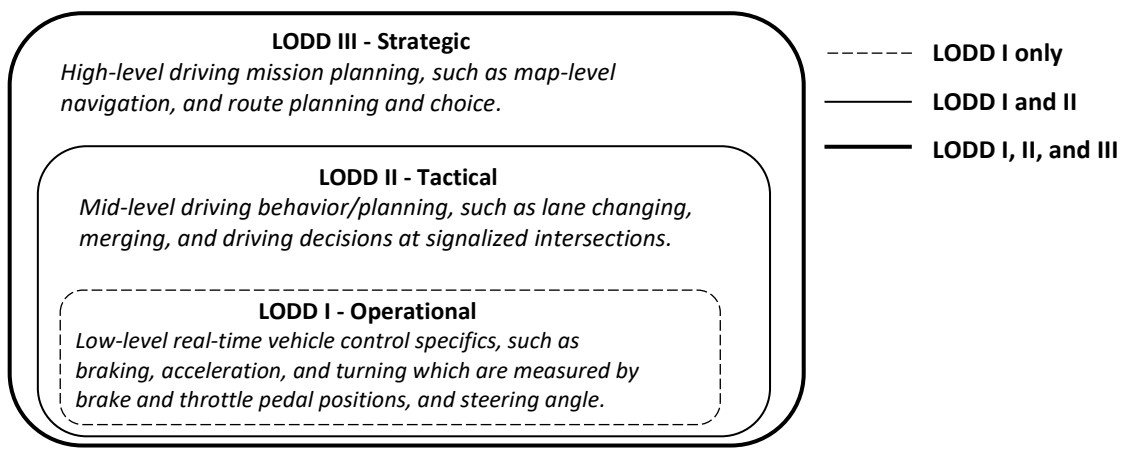

Figure 2. Levels of Driving Decisions

Table 1 presents a brief summary of some existing literature related to the levels of driving decision. This synopsis suggests that, at the current time, space-variant information acquired by the CAV (by virtue of its automated and connectivity features) have not yet been fused using advanced data-integration techniques, for CAV operations purposes. The efficient fusion of such data is important because given the complexity of the typical driving environment, there exists significant variation in the size of data input to be fed to the CAV controller at each time step. Such variation can be attributed to the variation in the number of vehicles in the CAV's vicinity within time intervals of duration as small as 1 second. In this case, the existing optimization-based AV control and planning algorithms may suffer implementation failure due to their inability to deal directly with such data input size variations that are encountered in real-life operations. For this reason, there is a critical need for highly representative preprocessors that can transfer the original varying-size inputs into fixed-size inputs. The complexities of driving tasks and typical driving environments make it rather difficult to manually design such preprocessors appropriately. Alternatively, DL could be used to extract useful representations from the raw data and to simultaneously develop an optimal control policy for the CAV. Therefore, DRL-based intelligent controllers can potentially offer elegant and effective solutions to the problem of dynamic input lengths of the incoming information and, ultimately, fuse spatio-weighted information to enhance the CAV's movement control. This is the primary motivation of the current paper. This motivation is rooted in the premise that if equipped with an appropriate reward function, a CAV driving algorithm can outperform human drivers because it is capable of making instantaneous and reliable driving decisions (J. Chen et al., 2019).

Table 1. A Synopsis of Past Research on DRL Applications in CAV Operations

| Application | Level of Decision | AI Technique | CAV context | References |
|---|---|---|---|---|
| Train the AV to learn smooth and efficient driving policy for lane change maneuvers | Tactical+Operational (LODD I & II) | Dueling Deep Q-network with Dueling Structure (DDQN) + Deep Q-network (DQN) | AV | (Qi et al., 2019) |
| Control the AV's movement in complex urban scenarios | Tactical+Operational (LODD I & II) | Double Deep Q-Network (DDQN) + Twin Delayed Deep Deterministic Policy Gradient (TD3) + Soft Actor Critic (SAC) | AV | (J. Chen et al., 2019) |
| Use V2V and V2I communication to derive data-driven driving policies for the AV | Tactical (LODD II) | Deep Imitation Learning + Long-Short Term Memory (LSTM) networks | CAV | (De Silva et al., 2018) |
| Establish the AV's optimal policies for overtaking/tailgating | Tactical (LODD II) | Maximum Entropy Inverse Reinforcement Learning (MEI-RL) | AV | (You et al., 2019) |
| Model the AV's decision on stop/go at a traffic light | Tactical (LODD II) | Hierarchical Policy Gradients (PG) | AV | (Chen et al., 2018) |
| Solve the "freezing robot" problem & execute successfully a safe and comfortable merge into dense traffic | Tactical (LODD II) | Model Predictive Control (MPC) + Proximal Policy Optimization (PPO) | AV | (Saxena et al., 2019) |
| Train the AV to learn smooth and efficient driving policy for lane change maneuvers | Operational (LODD I) | Deep Q-network (DQN) | AV | (Wang et al., 2018) |
| Train the AV to explore in a parking lot and avoid objects | Operational (LODD I) | Proximal Policy Optimization (PPO) | AV | (Folkers et al., 2019) |

In this paper, we focus on a LODD II decision, the CAV's lane-changing decisions because it has been found to be a prevalent crash-prone driving maneuver (Pande and Abdel-Aty, 2006; Sen et al., 2003; Sun and Elefteriadou, 2010; Zheng, 2014; Du et al., 2020b). Lane-changing can be described as a switch from one lane to another in a multi-lane traffic stream environment. Crashes attributed to lane-changing and related maneuvers constitute approximately 5% of reported roadway accidents and 7% of crash fatalities (Hou et al., 2015) . There exists a number of research studies on lane changing decision and the use of machine learning to enhance this operations maneuver. These include Hou et al. (2014) who used Bayes classifier and decision tree to predict lane-change decisions. Yang et al. (2018) modeled dynamic lane-change trajectory planning for AVs based on the states of neighboring vehicles. Suh et al. (2018) developed a model for lane-change trajectory planning with a combination of probabilistic and deterministic prediction based on an automated driving environment. Xie et al. (2019) used deep learning to model the lane-change process including the phases of decision-making and implementation of the lane change. Ali et al. (2018) found that a connected vehicle environment can enhance the efficiency and safety of mandatory lane changes. Zhang et al.

(2019) predicted lane-change behaviors based on vehicles' time-series features, using a deep learning model. Chen et al. (2020) applied a machine learning classifier to predict the lane-changing risk based on the space-series features of vehicles at the beginning of the lane-change maneuver.

**Research gaps**
In the CAV operations control literature, the existing body of research is dominated by models that involve the processing of fixed-size inputs. For example, Mirchevska et al. (2017) considered a model that process a maximum of twenty (20) features from vehicles in a CAV's vicinity. In addition, several researchers including Saxena et al. (2019) used occupancy grid representations with a fixed number of grids. In such grid representation, a Convolutional Neural Network (CNN) approach is needed to extract the useful information. These models are considered state-of-the-art; however, it can be argued that imposing a fixed size of the grid is unduly restrictive in terms of the number of vehicles in the CAV's environment that can be recognized, and therefore will be difficult to implement in the real world. Specifically, the fixed grid size approach has three shortcomings that are particularly debilitating. First, the grid size is incapable of changing to reflect changes in the CAV's connectivity range during the operations of the CAV. These changes in the connectivity range are realistic and are expected to occur in any real-world CAV operations environment. Second, it generally has a lower precision which can be attributed to so-called black box nature of neural networks. Third, it tends to be rather excessively expensive from a computational standpoint, and this precludes its effective integration for purposes of real-time decision making by CAVs.

To mitigate the limitations associated with fixed-size representation, Huegle et al. (2019) combined Deep Sets (Zaheer et al., 2017) and Q-learning (Watkins and Dayan, 1992) to address the variability in input size while the CAV algorithm "learns" to control the CAV. In essence, the Deep Sets procedure takes the totality of the embeddings and transfers them into fixed-sized features. The advantage of the resulting structure is the large flexibility in the number of inputs, which allows the model to create permutation invariant features. The sequence length of inputs does not affect the decision-making process. However, due to the Deep Sets's simple summation manipulation, the high dimension features are condensed into a single fixed-size vector, and useful information such as the speeds, locations, and lane positions of downstream vehicles, become lost in the process. Further, the absolute value of feature embeddings may grow linearly with the number of surrounding vehicles, because a normalization term is not provided. For example, the absolute value of feature embeddings associated with eighty (80) surrounding vehicles clearly exceed those associated with forty (40) surrounding vehicles. Therefore, the model lacks robustness and may not be transferable across training and testing scenarios. For example, the model trained with 80 vehicles surrounding the CAV is not directly transferable to one with 40 surrounding vehicles. Additionally, due to the permutation invariant characteristic of Deep sets, the difference in the value of information depending on the distance between the CAV and the vehicle source, is not explicitly considered. In other words, the information from vehicles located far from the CAV is assigned the same importance as information from vehicles located close to the CAV. However, it is reasonable to posit that in driving operations, information from closer vehicles should be assigned weights that exceed the weight of information from vehicles located far away. Therefore, the direct application of Deep Sets is problematic. For this reason, in this paper, we address this need for distance-variant value of information and we design a fusion method that appropriately assigns weights to the information received from the downstream vehicles based on their relative spatial locations.

Another research gap in Deep Set Q learning proposed by Huegle et al. (2019) is that their model is trained on datasets only with successful and safe lane-changing transitions;

unsuccessful maneuvers are not included in the training data. Therefore, their model does not punish movements that are inherently unsafe and therefore is unable to guarantee that its driving decisions will be collision free. To address this issue, the model must rely on an additional lower-level safety module to restrict the AV's behavior so that it refrains from unsafe driving decisions. However, due to the imperfections in sensors or actuators, this lower-level safety module may fail in certain scenarios and could lead to a crash of the CAV. The current paper is based on the premise that this issue can be mitigated particularly where the DRL model itself is capable of making collision-free decisions.

**Main contributions of this paper**
In addressing this issue in the context of CAV operations, this paper makes three main contributions. First, it develops a DRL-based model that integrates (using modified a Deep Sets procedure) information that is locally-obtained and system-wide information collected using connectivity capabilities of the vehicles. Secondly, the paper develops an end-to-end framework that uses the fused information to control the CAVs lane-changing decisions in a manner that minimizes collision risk. Thirdly, the paper assesses the effect of traffic density on the sufficiency of the connectivity range and provides an indication of the connectivity threshold to ensure desirable operational performance (in terms of travel efficiency, safety and comfort) of the CAV.

In this paper, we will show how these contributions reinforce the justification not only for having connectivity in prospective autonomous vehicles, but also for installing connectivity capabilities in existing human-driven vehicles particularly during the transition period when the traffic stream is shared by CAVs and connected HDVs. It is anticipated that such justification will resonate well in the realms of the state of practice and the state of the art. This is because transportation agencies have a fiduciary stake in ensuring road system efficiency, providing real-time information to road users, and monitoring performance of the taxpayer funded road infrastructure system. To these agencies, this results of this paper may provide motivation to establish policies that promote connectivity capabilities in HDVs and ultimately, realize these systemwide benefits. In offering this potential contribution, this paper hopefully provides a platform upon which stakeholders can realize the benefits of system connectivity to CAV operations, in terms of the CAV's operational efficiency and the optimal range of connectivity.

The remainder of this paper is organized as follows: The study methodology section describes the DRL basics, proposed method and the model architecture. The experiment settings section presents the DRL settings and the details of the implementation on a simulated test track. The results section compares our proposed model with other baseline models and uses a case study to identify the optimal connectivity range for a given set of traffic conditions. Also in this section, we demonstrate the practical limitations of the classic Deep Set Q learning method proposed by Huegle et al. (2019) in terms of model transferability across different scenarios of traffic density. In the final section, we summarize the contributions of this paper, present the limitations of the paper, and prescribe future work in this research area.

**Study Methodology**
Using reinforcement learning, an agent can explore the environment and subsequently learn a behavior that promotes desired outcomes and avoids undesired outcomes (Mousavi et al., 2018). In this process, the agent observing the current states, takes action, and receives feedback (a positive or negative reward) from the environment (which is the driving space, in the context of this paper). The agent evaluates the feedback signal, and understands the benefits (positive reward) of good actions and the (negative reward) of errant actions. In this paper, we use reinforcement learning to facilitate safe and efficient movements of the CAV within in a simulation environment (Figure 3).

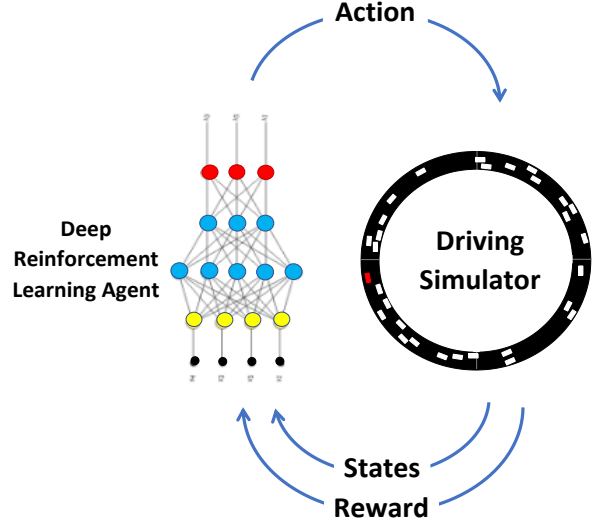

Figure 3. Reinforcement Learning in the Context of the CAV Driving Simulation

*Deep Q learning*
Time steps represent an essential feature of reinforcement learning processes in general. In a typical learning process, at each step, *t*, the learning agent undertakes an action, $a_t$, on the basis of (1) a policy network $\pi_\theta(a_t|s_t)$, which is parametrized as $\theta$, and (2) a current state of "nature", $s_t$. The agent carries out action $a_t$ and consequently enters a different state $s_{t+1}$ in accordance with the state transition distribution $p(s_{t+1}|s_t, a_t)$, and earns a reward $r_t$. Reinforcement learning seeks to learn an optimal policy network $\pi_{\theta^*}$ with $\theta^* = argmax_\theta \mathbb{E}[\sum_t r(s_t, a_t)]$. This enables the agent to earn a maximum sum of rewards between the time $t = 0$ to the time at the conclusion of the training episode. In this paper, we adopt broadly, the Q learning, a model-free method for purposes of identifying the optimal driving policy. The Q function $Q^\pi(s_t, a_t) = \mathbb{E}_{a_{t'>t} \sim \pi_\theta}[r(s_t)|a_t, s_t]$ is a representation of the total expected reward from time *t* after choosing the action $a_t$, over the entire trajectory. The Q function not only provides an easy way to evaluate how "good" the choice of $a_t$ value is, but also gives guidance on the choice of a driving policy that yields a maximum value of the Q function. Recognizing the inherent difficulty of expressing the Q function in an explicit manner, we use a deep neural network technique to yield an approximation of the Q function (this is termed a classical Deep Q Network (DQN) method, which was also applied in (J. Chen et al., 2019; Qi et al., 2019; Wang et al., 2018). We also use a replay buffer not only to increase the robustness of the model in all the situations but also to avoid overfitting issues associated with certain problem scenarios. This is a much needed step where it is sought to generate random experiences for training.

*Overview of the model*
With regard to the input space of the model, we consider explicitly at each time step *t*, 3 blocks of state. This includes the information from downstream sources (out of the sensing range but within the connectivity range) $X_d$; information from proximal or "local" sources (that is, information sources that are within the range of the CAV's sensors), $X_l$; and the CAV's information, $X_{CAV}$. In sum, the overall state space can be represented as a triplet $(X_d, X_l, X_{CAV})$. Information from the farther (downstream) sources are characterized as being of "variable length", that is, it changes when there is a change in the number of vehicles in the CAV's within its connectivity range. To address this variable length input problem, we adopt in this paper, a similar Deep Set concept to aggregate the dynamic sized input into a fixed shape but with a superior normalization mechanism. The second information source captures the driving environment within the close neighborhood of CAV, which is incorporated to promote collision-free decisions by the CAV. The inherent large amount of detail is needed to fully describe the movement attributes of vehicles located in the same lane as the CAV, and those located on the lanes left and right of the CAV. In this paper, we divide further, the local inputs into "left" lane, "right" lane and the "current" lane (the current lane is that which is occupied by the CAV). Information from the third source (that is, from the CAV itself), which includes its absolute location, speed and lane position, is provided as the final block of inputs to the CAV control system.

In Deep Sets, variable lengths of inputs are first fed into an encoding network to gain proper feature embeddings separately for each input. In this paper, we adopted this concept to use fully connected neural networks $\varphi$ to encode each downstream vehicle input $x_d \in X_d$ within the connectivity range, the input from each sensed lane $x_l \in X_l$ within the vicinity of the CAV, and the CAV's information $x_{CAV} \in X_{CAV}$ into a higher dimension feature space. Then we perform information fusion for the dynamic changing length among the feature space. Here, we simply use the same encoding network for both downstream and local inputs because they have the same meaning and representations. After the encoding network, the downstream embeddings are weighted and summed to obtain a fixed size input for subsequent operation. The total feature embedding obtained from downstream information is:

$$F_d = \sum_{i=1}^{n} w_i \varphi(x_d^i)$$

Where: $x_d^i$ and $w_i$ the raw feature input and weight for $i^{th}$ vehicle that is located downstream of the CAV. The weight values represent the relative importance of information from the various sources, for the CAV driving purposes, and the sum of weights of information from all vehicles in the connectivity range is 1.

The local information sources are: "left", "right" and the "current" lanes. A matrix can be used to represent the feature embedding which contains information associated with these 3 lanes, as follows:

$$F_l = \begin{pmatrix} \varphi(x_l^{left}) \\ \varphi(x_l^{current}) \\ \varphi(x_l^{right}) \end{pmatrix}$$

The embeddings of CAV's information has a similar expression as follows:

$$F_{CAV} = \varphi(x_{CAV})$$

The model concatenates the feature embeddings for downstream, local and CAV information to yield a fixed-sized feature map. Then the feature map is flattened and fed into the Q network $\rho$ for Q values. Denoting the overall model that contains the encoding network and Q network as: $\hat{Q}$, with parameters $\theta$, the final Q values can be expressed as:

$$\hat{Q}_\theta(s_t, a_t) = \rho([F_d^t; F_l^t; F_{CAV}^t], a_t)$$

The encoding network and Q network are trained on mini-batches sampled from a replay buffer R, which contains the transitions of $(s_t, a_t, r_t, s_{t+1})$. For each mini-batch, the objective of the training is to minimize the following loss function:

$$L_\theta = \frac{1}{b}\sum_t y_t - \hat{Q}_\theta(s_t, a_t)$$

Where: $b$ is the batch size and $y_t = r_t + \gamma \max_a \hat{Q}_\theta(s_{t+1}, a)$. In Figure 4, we present the layout of the model. Multi-layer perceptron (MLP) is used for each component with the following architecture:

- Encoding network $\varphi$ : $Dense(64) + Dense(32)$
- Q network $\rho$: $3 \times Dense(64) + Dense(32) + Dense(16) + Dense(8)$
- Output layer: $Dense(3)$

It is sought to facilitate full exploration (by the agent) of the environment and to acquire adequate experiences in both categories of driving success and failure (collision). Therefore, we use a Deep-Set Q learning that incorporates an experience reply buffer and a "warming up" phase with total T steps that allows the agent to undertake random actions. From step T+1, we perform training by maximizing the reward and minimizing the losses, as mentioned above. To further reduce the variance for the model, we apply a double Q learning mechanism with a soft updating for target network as introduced in (Van Hasselt et al., 2016). Algorithm 1 (below) presents the steps for the entire process.

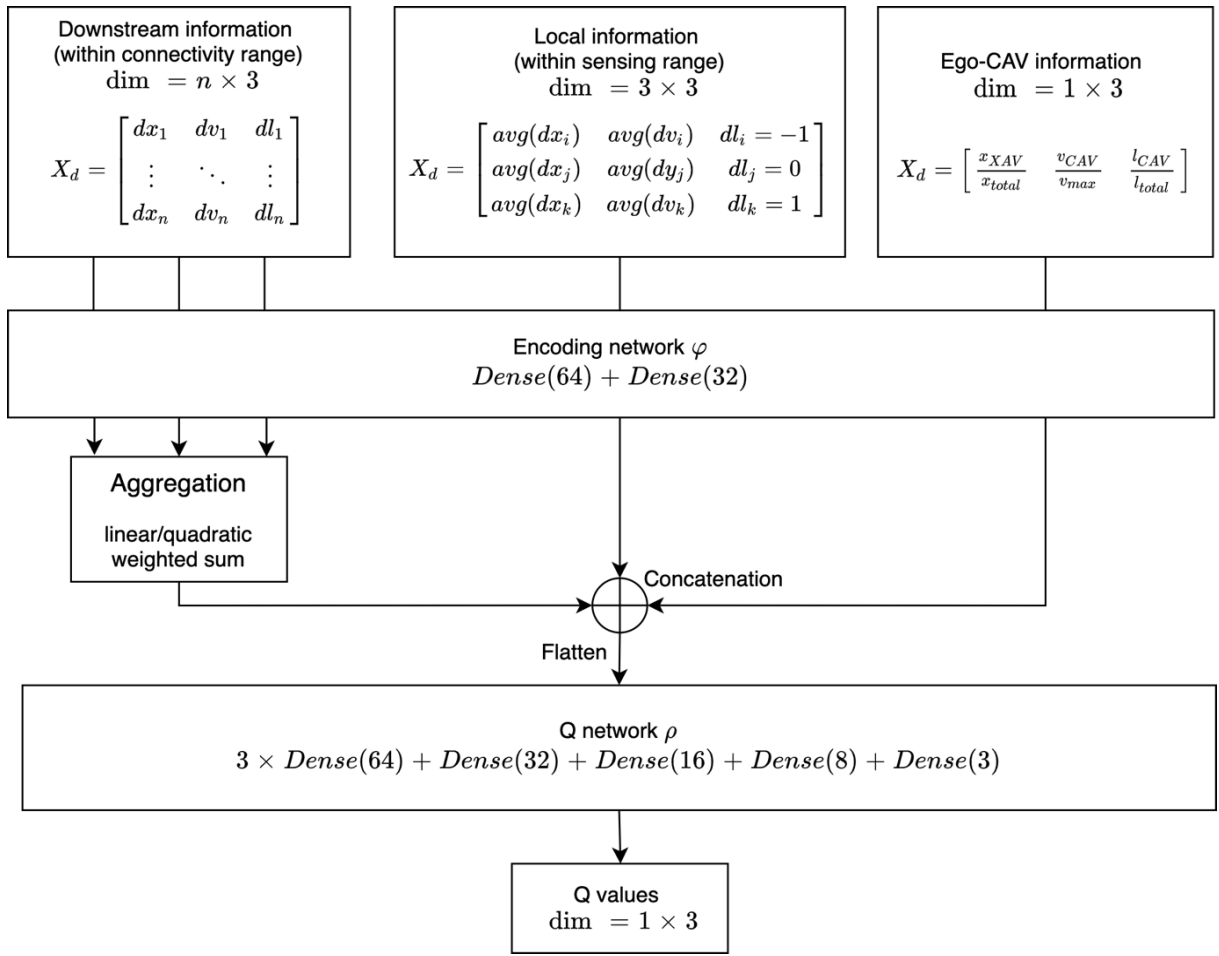

Figure 4. Proposed network architecture

| Algorithm 1 | Spatially Weighted Deep-Set Q Learning with Experience Replay and Target Network |
|---|---|

Initialize the reply memory $R$ to capacity $N$
Initialize the weights for both Encoding network $\varphi$ and Q network $\rho$ which jointly denoted as Network $\hat{Q}_\theta$ and Target Network $\hat{Q}_t = \hat{Q}_\theta$
## Warming up steps
For time step $t = 1$ to $T_1$ (warming up steps) **do**
    Take a random action $a_t = a_r$ and gather the transition $(s_t, a_t, r_t, s_{t+1})$
    Store the transition $(s_t, a_t, r_t, s_{t+1})$ into the memory buffer $R$
## Main training loop
For time step $t = T_1 + 1$ to $T$ (training steps) **do**
    ## Generate new samples and update memory R
    With probability $\epsilon$ select a random policy $a_t = a_r$
    Otherwise:
    Encode the information from the CAV directly, downstream sources and sources in the immediate locality, with $\varphi$ and weights $w_i$

$$F_d = \sum_{i=1}^n w_i \varphi(x_d^i), \quad F_l = \begin{pmatrix} \varphi(x_l^{left}) \\ \varphi(x_l^{current}) \\ \varphi(x_l^{right}) \end{pmatrix}, F_{CAV} = \varphi(x_{CAV})$$

    Obtain action $a_t^* = \underset{a_t}{\mathrm{argmax}}\, \hat{Q}_\theta(s_t, a) = \underset{a}{\mathrm{argmax}}\, \rho([F_d^t; F_l^t; F_{CAV}^t], a_t)$
    Execute $a_t^*$ and observe reward $r_t$ and next state $s_{t+1}$
    Store transition $(s_t, a_t^*, r_t, s_{t+1})$ into the memory buffer $R$
    Set $s_t = s_{t+1}$
    ## Training the model at each training step
    Sample random mini-batch with size b from R
    For each training examples with the batch, set the target of Q value

$$y_t = \begin{cases} r_t + \gamma \max_{a_{t+1}} \hat{Q}_\theta(s_{t+1}, a_{t+1}) & \text{if } s_{t+1} \text{ is not done} \\ r_t & \text{if } s_{t+1} \text{ done} \end{cases}$$

    Perform a gradient step optimizing loss function in $L_\theta = \frac{1}{b}\sum_t y_t - \hat{Q}_t(s_t, a_t)$
    ## Updating the Target Network
    If mod(t, target updating frequency) == 0
        Set $\hat{Q}_t = \hat{Q}_\theta$

**Experiment settings**

In this paper, we use an open-source simulator SUMO (Krajzewicz et al., 2012) with a Python library "flow" developed by Kheterpal et al., (2018) to create the RL environment and to control the autonomous vehicles. Also, we use SUMO to ultimately render and visualize the model behavior. As the proposed DRL seeks to model a Markov Decision Processes (MDP), we first define the state space, action space and the corresponding reward function.

*State space*
As we state in an earlier section of the paper, we divide our state space into 3 parts (Figure 5): the local information part (obtained through the sensors), the downstream information part (obtained from the CAV's connectivity capabilities), and the movement information of the CAV itself.

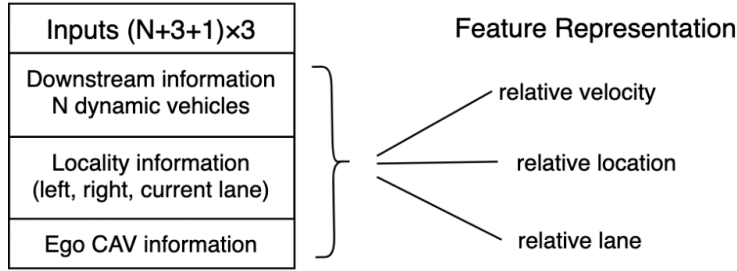

Figure 5. Representation of tripartite state space

The connectivity related information from downstream sources is acquired within a connectivity range that is denoted by $dx_{connectivity}$, locality refers to the area immediate surrounding the CAV within a sensing range of $dx_{local}$. For both local information and downstream information, we consider the same representation including the following three features, relative distance, relative speed, and relative lane position defined as follows:

Relative distance: $dx_i = \frac{x_i - x_{CAV}}{dx_{connectivity}}$

Relative speed: $dv_i = \frac{v_i - v_{CAV}}{v_{max}}$

Relative lane: $dl_i = l_i - l_{CAV}$

Where:

$x_i$ is the absolute position of $i^{th}$ vehicle that is in front of the CAV and $x_{CAV}$ is the CAV's absolute position, $x_i$ and $x_{CAV}$ are both with respect to the global coordinate frame.

$v_i$ and $v_{CAV}$ are the speed of $i^{th}$ vehicle that is immediately leading the CAV and the CAV's speed, respectively, $v_{max}$ is the normalization term. In the given scenario, we use the road speed limit to represent the operating speed.

$l_i$ and $l_{CAV}$ are the lane index of $i^{th}$ vehicle and the CAV, respectively.

Therefore, it is possible to construct a representation of the downstream information using the matrix $X_d$ below:

$$X_d = \begin{pmatrix} dx_1 & dv_1 & dl_1 \\ \vdots & \ddots & \vdots \\ dx_n & dv_n & dl_n \end{pmatrix} \in \mathbb{R}^{n \times 3}$$

Where: $n$ is the number of vehicles that are within the connectivity range.

Regarding the local information (that is, information from the sensors), we recognize that the number of vehicles within the sensor range is not constant. Therefore, we perform an "averaging" manipulation in order to render the local information on the same scale as that of a single vehicle.

$left = (avg(dx_i) \quad avg(dv_i) \quad dl_i = -1)$
$right = (avg(dx_i) \quad avg(dv_i) \quad dl_i = 1)$
$current = (avg(dx_i) \quad avg(dv_i) \quad dl_i = 0)$

$$Locality = \begin{pmatrix} left \\ current \\ right \end{pmatrix} \in \mathbb{R}^{3 \times 3}$$

With regard to the CAV information, we consider the "relative" quantity with respect to the total quantity in the scenario, specifically, the relative distance is with respect to the total length of the track; the relative speed is with respect to the speed limit; and the relative lane is with respect to the total number of lanes on the track.

$$CAV_{info} = (\frac{x_{CAV}}{x_{total}} \quad \frac{v_{CAV}}{v_{max}} \quad \frac{l_i}{l_{total}})$$

*Weights*
We define our model in a linear weighted manner, so that we can assign weights explicitly to the dynamic inputs. For each HDV, the feature embedding is inversely proportional to its relative distance to the CAV. The sum of weights is 1.

$$w_i = \frac{1/dl_i}{\sum_i^n 1/dl_i}$$

We also develop another quadratic weighted baseline model for purposes of comparison, and we define the weights as inversely proportional to the square of the relative distance, as follows:

$$w_i = \frac{1/dl_i^2}{\sum_i^n 1/dl_i^2}$$

*Action space*
For each time step, the action space is discrete and represents the potential actions to be undertaken by the CAV, as follows:
$\mathcal{A} = \{change\ to\ left, keep\ lane, change\ to\ right\}$.
The simulator restricts the agent from exiting the simulated corridor.

*Reward function*
In this paper, we use two types of rewards and two types of penalties: the destination reward, speed reward, lane-changing penalty, and collision penalty, as listed in Table 2.

Table 2. Rewards and Penalties

| Category | Description | Symbol |
|---|---|---|
| Speed reward | Defined as the relative speed w.r.t speed limit, representing travel efficiency. | $R_v = \frac{v_{CAV}}{v_{max}}$ |
| Destination reward | An instant reward assigned to the agent when it accomplishes the ultimate goal by reaching a specified "final" destination. | $R_D$ |
| Collision penalty | A fixed value that discourages maneuvers which may lead to a crash. This penalty is typically assigned a large value to encourage the model to learn safe decisions but not too large as that would lead to extremely conservative driving and low travel efficiency. | $P_c$ |
| Lane changing penalty | Recognizes the inherent crash risks associated with lane changing. Used to discourage the agent from excessive lane-changing maneuvers. | $P_{LC}$ |

The overall reward function is defined as:
$R_{total} = w_1 R_v + w_2 R_D - w_3 P_c - w_4 P_{LC}$
Where:
$w_1 \ldots w_4$ are weights that can be tuned by the model while duly recognizing the trade-off between the "ride comfort" and "travel speed". When $w_1$ and $w_2$ are tuned at relatively higher levels compared to $w_3$ and $w_4$, the speed reward ($R_v$) term and destination reward $R_D$ term will dominate and this will encourage the vehicle to perform maneuvers that are associated with high speed and aggressive driving, and therefore will reach the destination as fast as it can. On the other hand, an increase $w_3$ and $w_4$ relative to $w_1$ and $w_2$ means higher penalties for unsafe and lane changing behavior, in which case the model eschews aggressive behavior including frequent lane changing in favor of lane keeping (that is, less frequent lane changing), thereby avoiding crash prone maneuvers. In this case, the model learns to become more conservative and safety-conscious and reaches the destination in a time that is longer compared to the previous case.

*Simulator parameters*

The driving simulation environment is defined by the parameters used in the simulator. Therefore, the simulation scenarios differ from each other in terms of the following attributes: overall traffic network features, the vehicle control logic which defines the interaction between vehicles, and other environment settings made specifically for training the AI, as discussed below.

(i) Scenario parameters

In the experiments described in this paper, we used a 500-meter, 4-lane circular loop track. For the AI training and evaluating all the experiments, we adopted the following mixed-traffic specification (MTS): (51; 2%; 98%), that is, 51 vehicles comprising 1 CAV and 50 HDVs. In the initial step, HDVs were generated at locations spaced out uniformly on the road track. To lend heterogeneity to the experimental setting, the HDVs were introduced with a random initial speed ranging from 0 to 15 m/s and a random maximum speed from 15 to 30m/s. The road segment speed limit is set as 50m/s. In the experimental setting, this speed limit can be reached only by the CAV.

(ii) Training environment parameters

For each training episode, the only termination criterion is to reach the maximum time step of 1200, to allow the CAV to travel 3-4 rounds in the loop. Therefore, if the speed is adequately high, the agent will earn both higher speed reward and more destination reward by completing a greater number of rounds.

(iii) Vehicle control parameters

The vehicles on the road track (both HDVs and the CAV) were controlled longitudinally in acceleration using Trieber and Kesting's Intelligent Driver Model (Treiber and Kesting, 2013), a built-in controller SUMO simulator. For HDVs, an additional normal noise with 0 mean and a random standard deviation from 0-1 is added to characterize further, the acceleration of each HDV, to replicate the unpredictable driving nature of human drivers, and to ultimately produce stop-and-go traffic conditions on the road track. A rule-based baseline model was used for the transverse control; this is a strategic lane-change concept. Here, we use the SUMO built-in lane changing model LC 2013 (Erdmann, 2015) to control HDVs and the rule-based CAV, and uses the model's default parameters. In the experimental setting, we set the CAV such that it takes two (2) seconds to execute a complete lane-change operation.

*Training parameters*

The deep learning model was trained online using a reply buffer. First, the reply buffer was filled with $5 \times 10^5$ random warm-up transitions before the start of the training. After the training started, the transition batches with $batchsize = 32$ were sampled randomly without replacement and fed to the model. Then the buffer was updated with the new transitions. The overall training horizon, including warm-up and actual training, was established at $10^6$ steps (i.e., approximately 833 epochs) in total. To analyze the tradeoff between exploration and exploitation, we use a simple epsilon greedy policy with a probability of 0.3 for exploration and 0.7 for exploitation. For the optimization parameters, we use *Adam*, a method for stochastic optimization (Kingma and Ba, 2015) with an initial learning rate of $\gamma = 10^{-4}$ and a soft target model update rate $\tau = 10^{-2}$.

**Results**

*Training process*

In the training process (Figure 6), the first $5 \times 10^5$ steps (417 epochs) are "warming up" phase that indicates the reward for making random choices. This phase is intended to equip the agent with a sufficient learning experience that contains both successes and failures. The training commences after $5 \times 10^5$ steps and converges in $10^6$ steps (833 epochs). Specifically, the "jump" at approximately 420 epochs is a gradual increase which goes up along with the training process. In our case, the model converges fast compare to the "warming up" phase and the convergence phase. After the training, the CAV can perform lane changing maneuvers without collision.

*Comparative analysis*

We compared the results from our proposed model with the four baseline operation decision models: the unweighted Deep Set Q learning model, the quadratic weighted Deep Set Q learning model, the rule-based lane-change model, and the no-lane-change model. The mean and median performance are compared in Figure 7 and Table 2.

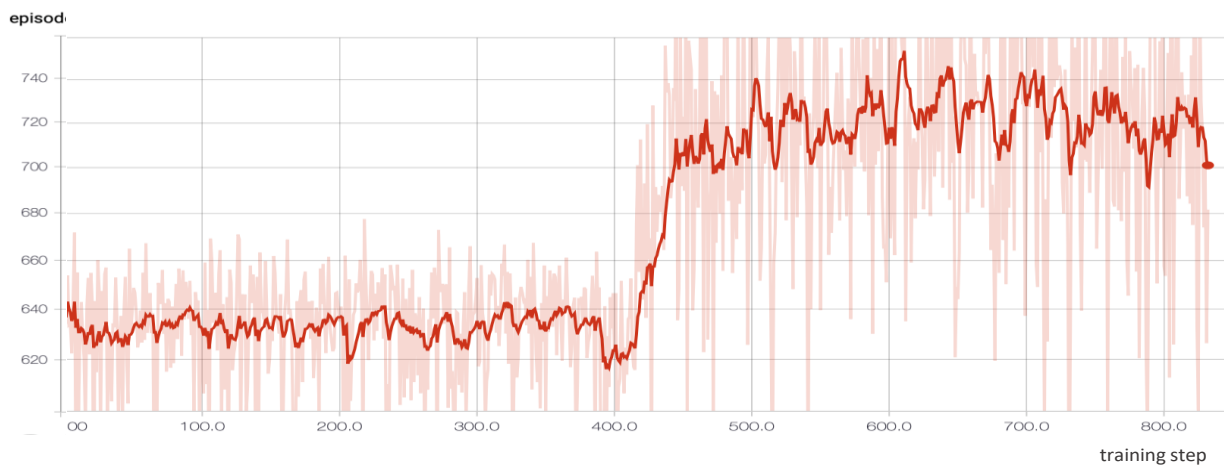

Figure 6. Rewards gained vs. the number of training steps

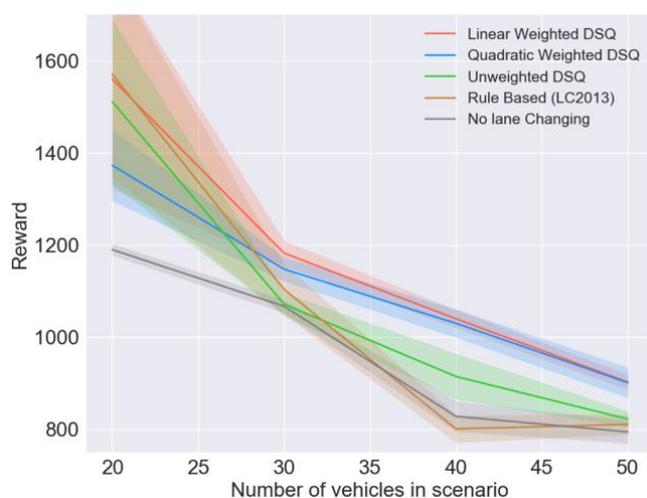

(a) Mean with 95% confidence interval

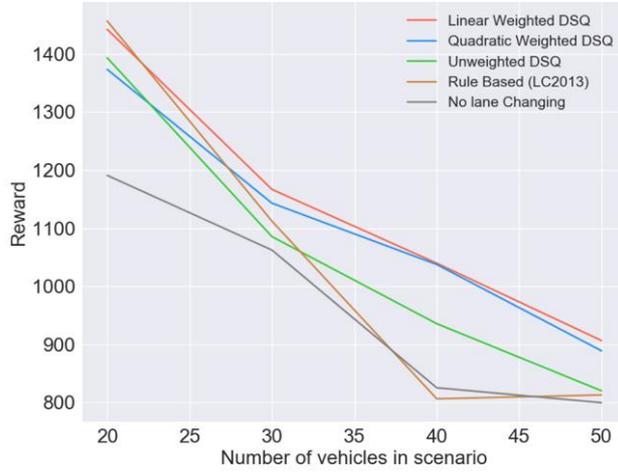

(b) Median

Figure 7. Relative performance of the models investigated, using 10 test episodes

Table 2 Performance comparison for different models in different scenarios.

| Scenarios | Models | No Lane Changing | Rule Based (LC 2013) | Unweighted DSQ | Quadratic Weighted DSQ | Linear Weighted DSQ |
|---|---|---|---|---|---|---|
| 20 vehicles | mean | 1189.88 | **1570.49** | 1510.71 | 1372.78 | 1559.57 |
| | median | 1191.04 | **1456.63** | 1393.34 | 1373.38 | 1442.28 |
| | S.D. | **19.47** | 359.22 | 287.22 | 123.72 | 382.47 |
| 30 vehicles | mean | 1066.95 | 1103.65 | 1071.63 | 1147.56 | **1182.19** |
| | median | 1062.43 | 1112.17 | 1085.52 | 1143.11 | **1166.82** |
| | S.D. | **21.81** | 87.45 | 38.47 | 40.81 | 42.26 |
| 40 vehicles | mean | 828.18 | 801.07 | 914.74 | 1030.15 | **1039.95** |
| | median | 825.5 | 806.57 | 935.76 | 1037.64 | **1039.87** |
| | S.D. | 51.99 | 47.13 | 79.23 | 48.78 | **30.57** |
| 50 vehicles | mean | 794.64 | 810.93 | 822.45 | 901.41 | **902.7** |
| | median | 799.86 | 813.15 | 820.27 | 889.09 | **906.68** |
| | S.D. | 38.73 | 34.38 | 26.34 | 51.94 | **25.44** |

The results were tested in various scenarios that differ in terms of their traffic densities. From Figure 7, it can be seen that in most scenarios, the linear weighted CAV decision model outperforms the unweighted and the quadratic weighted models, and all these three are superior to the no-lane-change and rule-based baseline decision models. In one scenario with low traffic density (that is, only 21 HDVs on the road track) and where traffic conditions approach free flow, it was found that all the HDVs and the CAV operated at speeds that approach their maximum possible speed under such stable traffic conditions. Also in this traffic density scenario, for each of the five models, the CAV algorithm was found to make a consistent decision, that is, the CAV keeps in its lane. This is intuitive because under such traffic conditions, there is no incentive for the CAV to change lanes. Further, in this scenario, it is observed that the slower HDVs stay in the rightmost lane and leave the left lane to the other vehicles that have higher speeds. When the rule-based decision model is used in this scenario, we observe that the CAV "captures" the leftmost lane where it maintains a high speed. In that scenario, the rule-based model provides the CAV the highest reward compared to other models. In another traffic density scenario that involved all 51 vehicles, it is observed that at such high traffic density, the vehicles cannot gain much travel benefit even after making lane changes. In that scenario therefore, all the models were found to yield similar reward level. In a third scenario with traffic density that is in-between the first two scenarios described earlier (that is, 30-40 HDVs on the road track), we observe that the CAV can greatly enhance its operational efficiency by making appropriate lane-changing decisions as and when needed. We find that in this scenario, our proposed 2 "weighted DSQ" models outperform the other 3 baselines while the linear weighted model is slightly superior to the quadratic weighted model. This result may be attributed to the model's capability to obtain and appropriately process (through weighting), the information on traffic conditions further downstream (due to its connectivity capabilities) and traffic conditions in its immediate local environment (due to its sensing capabilities). This capability helps it to identify an optimal driving policy under the given traffic conditions, and to make proactive decisions to avoid travel delay caused to it due to proximal or anticipated imminent delay threats in the traffic environment.

*Optimal connectivity range*
In the context of this research, connectivity range refers to the maximum distance at which connectivity is available. Therefore, the "optimal" connectivity range refers to distance after which any additional benefits of increasing connectivity are negligible. As we stated earlier in this paper, the developed model is capable not only of normalizing explicitly, the input scale but also of accounting for the spatial distribution of the inputs. Therefore, it is possible to use the same model under various specified connectivity ranges without the need to retrain the model. The results of the experiment (Figure 8) demonstrate that for a given traffic density, as the connectivity range increases, the model performance increases sharply up to a certain point after which it increases at a reduced rate and almost flattens out. This is because, when connectivity range in low, a unit increase in the level of this attribute causes a proportionately higher amount of downstream information to be sent to and received by the CAV. To the CAV's decision processor that seeks to make proactive decisions, the incremental benefit of such information is significant. However, when the connectivity range is large, a unit increase in the connectivity range will produce relatively smaller benefits. This is due to the increased variance arising from noise or unrelated information that is received by the CAV, a situation that is exacerbated by the unpredictable and often errant nature of human drivers in HDVs located further away from the CAV. This trend suggests the existence of an optimal connectivity range, in other words, a threshold beyond which the marginal benefits of increased range, begin to diminish. In this paper, we determine this threshold from the derivative of the trendline, which in general, is an indicator of this marginal benefit. In each scenario (21, 31, 41, 51 vehicles), we evaluate the derivative of the trendline at 100-meter connectivity range $x_0 = 100m$ as the baseline marginal effect $g_0$. We keep increasing connectivity range $x$ until the derivative of the trendline drops to $0.1g_0$, and then we observe that the marginal benefit drops to only 10% of baseline value, and the corresponding $x$ is the optimal connectivity range $x_{optimal}$. Based on our experimental settings, for all 4 scenarios, the optimal connectivity range $x_{optimal}$ is approximately 270m (Figure 9).

Further, in the scenarios with sparse traffic (21 vehicles in the corridor, that is, 20+1) and very dense traffic (51 vehicles), the convergence of reward is faster due to adding more information can barely improve driving in these scenarios. This can provide CAV manufacturers, the justification for specific optimal connectivity range specifications and for them to provide CAV users with flexibility to select appropriate optimal range under a given set of traffic conditions. In other words, to achieve high efficiency in information transmission and usage for its efficient operations, the CAV should be able to automatically identify and adopt a specific connectivity range setting or mode based on the prevailing traffic density it has sensed.

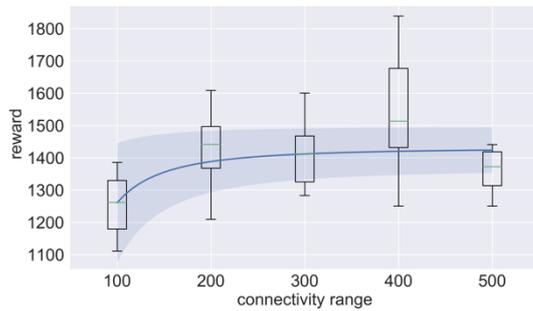

(a) Traffic density = 20

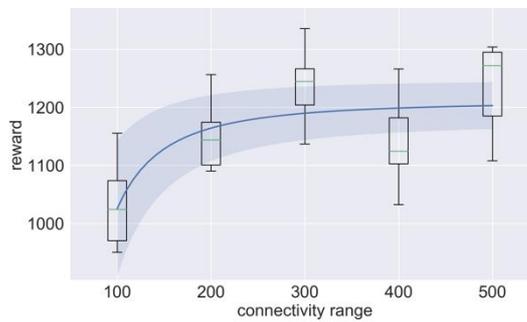

(b) Traffic density = 30

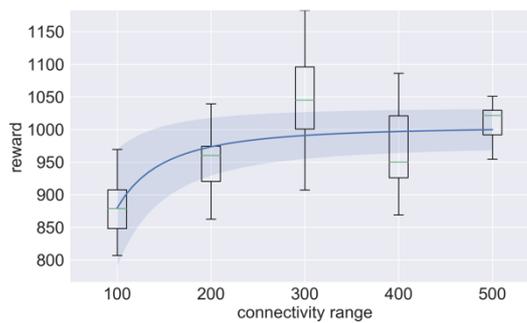

(c) Traffic density = 40

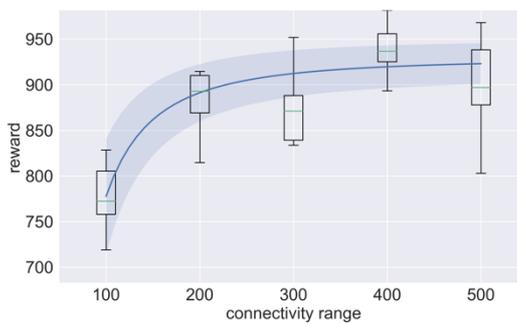

(d) Traffic density = 50

Figure 8 Reward vs. connectivity range, using the normalization manipulation (linear weighted DSQ) model

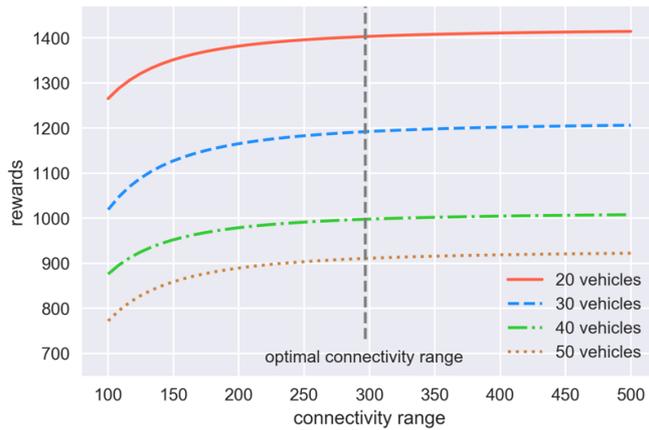

Figure 9 Effect of traffic density of the reward-connectivity range relationship

On the basis of the reward that is based on connectivity range, the marginal effect of increasing connectivity range on the reward was plotted (Figure 10). The figure shows that further increases of the connectivity range is not always beneficial because it exhibits fast diminishing returns in terms of the reward. This is seen for all four scenarios of traffic density, and the convergence of the curves representing the various traffic densities, seems to occur at approximately 270 m. The elbow points of the curves seem to be in the range 170-180 ft. The results can serve as a guideline for manufacturers of connected vehicle technology regarding not only the default setting of the connectivity range but also the manufacturer's recommended (and subsequently, CAV driver-adjusted) setting of the appropriate connectivity range setting for the prevailing traffic conditions (density). In some cases, a higher connectivity range may come at a higher cost to the driver. In such cases, both the marginal benefits and marginal costs of increased connectivity range will need to be considered in order to establish the most cost-effective level of connectivity, under a prevailing set of traffic conditions.

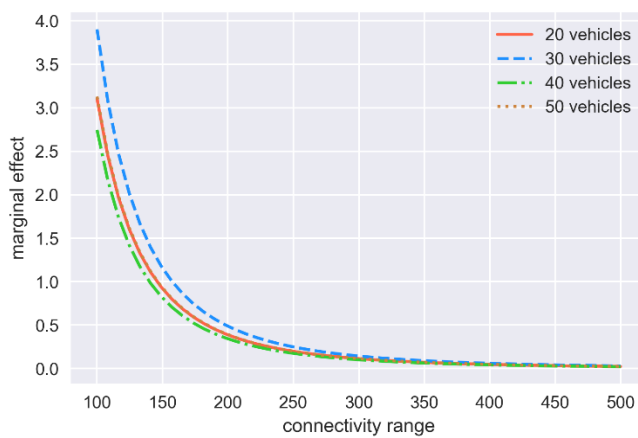

Figure 10 Marginal effect vs. connectivity range

**Analysis involving classic DSQ model**
As we discussed earlier in the "Research Gap" section of this paper, the unweighted Deep Set Q (DSQ) model proposed in (Huegle et al., 2019) may suffer from the problem of non-transferability across different traffic density conditions, unlike the normalization manipulation (weighted sum operation) model developed in this paper. To investigate this hypothesis, we perform the connectivity range experiment using the baseline unweighted DSQ model. The results are presented in Figure 11. As shown in the figure, for all the different traffic density scenarios, an increase the connectivity range does not lead to an improvement in CAV's performance, unlike Figure 8 (linear weighted DSQ). This is because for the unweighted DSQ, there is no proper normalization mechanism. The embedding scale of downstream information grows linearly with the number of connected vehicles in a fixed space. That is, the scenario with 80 vehicles has larger scale of feature input than that with 40 vehicles. Therefore, increments in the connectivity range will create an unbalance in scale between downstream information, local information and CAV information. When the connectivity range is very large, the unweighted DSQ model causes the downstream information to overwhelm the local information. However, local information is vital for some close-space maneuvers including lane changing. Therefore, in the unweighted DSQ model, such "wiping out" of the local information will lead to a drastic increase in crashes. For this reason, if the DSQ model is used, increases the connectivity range will generally not be seen to improve the CAV's performance, which is counter-intuitive. Therefore, the normalization manipulation (weighted sum operation) model in our proposed framework is more effective in accounting for the benefits of increased connectivity (without sacrificing the local information) and therefore is more appropriate for robust CAV operations.

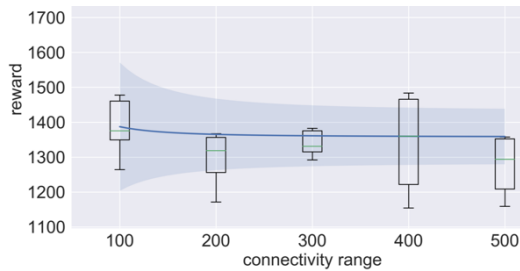

(a) Traffic density = 20

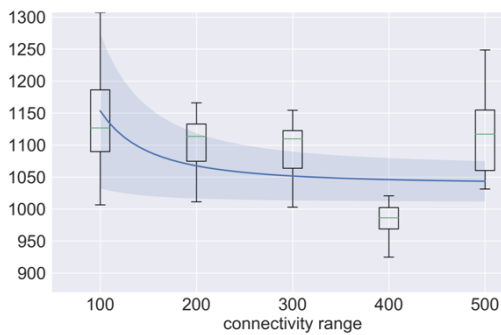

(b) Traffic density = 30

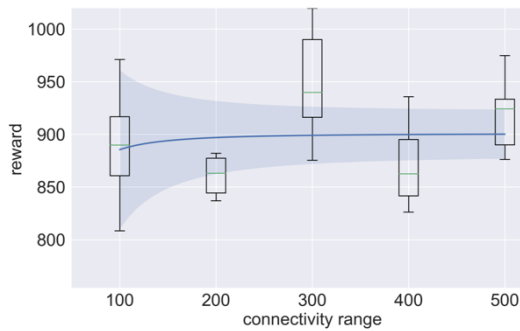

(c) Traffic density = 40

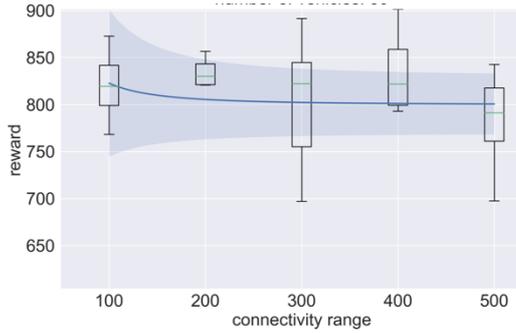

(d) Traffic density = 50

Figure 11 Reward vs. connectivity range, for different scenarios of traffic density, using the baseline (unweighted DSQ) algorithm

**Concluding remarks**
In this paper, we present an end-to-end deep reinforcement learning based processor to make high-level decisions in controlling a CAV's lane change operations in complex mixed traffic. In this context of operations, the developed model was observed to achieve its target of helping a CAV increase its travel effectiveness and efficiency in terms of safety and mobility, respectively. As part of efforts to achieve this overarching objective, the paper also demonstrates the efficacy of the proposed model in four areas. First, the model adequately fuses the long-range and short-range information based on the spatial importance of information which, in turn, in a function of the spatial distance between the information source and the CAV. Second, the model helps the CAV make safe lane-change decisions even after relaxing the collision-free restriction imposed by the low-level controller in the simulator. Third, the model handles adequately, the highly dynamic length of inputs (that is fed to the CAV). Finally, the model efficacy is demonstrated by applying it to traffic scenarios with different densities without the need to retrain the model. For a comparative evaluation, we compare the proposed model with four classic baseline methods (unweighted classic DeepSet Q learning method, quadratic weighted DeepSet Q learning method and the multiple-rule-based). The results suggest that the model proposed in this paper outperforms the baseline and other models.

With regard to the issue of connectivity range and issues of practical implementation, the paper demonstrates how the optimal connectivity ranges at a prevailing level of traffic density, could be ascertained. Therefore, the paper provides CAV manufacturers a justification for specific optimal connectivity range specifications. In addition, with the developed model, the CAV can automatically identify and adopt a specific connectivity range setting or mode based on the prevailing traffic density. Therefore, the model also presents to manufacturers, a capability to provide CAV users with flexibility to select appropriate optimal range under a given set of traffic conditions. In general, CAV manufacturers may find this useful in their efforts to develop appropriate vehicle connectivity protocols and architectures.

Moving forward to future work, with the help of connectivity and storage system, research may find it worthwhile to consider temporal information including historical data on the vehicle position, speed, and acceleration accounting for the possibility of longer times (delays) of the CAV's decision process. The incorporation of such historical data in the analysis may help address hypotheses regarding the effect of imminent traffic conditions downstream that often require rerouting or preemptive evasive maneuvers of the CAV. Examples of these downstream conditions include construction sites or workzones, accidents, debris, potholes, and obstacles on the roadway. Therefore, future research could examine the efficacy of trajectory planning in CAV by incorporating both instant (short-term) and long-term information. Also, future research could investigate the efficacy of DRL based method, for purposes of CAV control, in making collaborative decisions that maximize the utility of all agents in the entire corridor rather than the CAV's utility. An example of such research directions is the use of the proposed methodology to promote traffic string stability and cooperative crash avoidance maneuvers in emergency situations. Finally, in determining the optimal connectivity range, future studies may consider not only the marginal benefits as done in this paper, but a combination of both marginal benefits and marginal costs of connectivity range increments. The cost aspects could include the initial purchase/installation cost and operations cost of connectivity devices, and the cost of computing power to process the information obtained through connectivity.

**Acknowledgements**
This study is based on research supported by the Center for Connected and Automated Transportation (CCAT), Region V University Transportation Center funded by the U.S. Department of Transportation, Award #69A3551747105. An abridged version of the paper was submitted to the IEEE ITSC for presentation at the September 2020 IEEE conference in Greece.

Res. Part C Emerg. Technol. https://doi.org/10.1016/j.trc.2017.02.024

Qi, X., Luo, Y., Wu, G., Boriboonsomsin, K., Barth, M., 2019. Deep reinforcement learning enabled self-learning control for energy efficient driving. Transp. Res. Part C Emerg. Technol. https://doi.org/10.1016/j.trc.2018.12.018

Roberts, C.A., Attoh-Okine, N.O., 1998. A Comparative Analysis of Two Artificial Neural Networks Using Pavement Performance Prediction. Comput. Civ. Infrastruct. Eng. https://doi.org/10.1111/0885-9507.00112

Saxena, D.M., Bae, S., Nakhaei, A., Fujimura, K., Likhachev, M., 2019. Driving in Dense Traffic with Model-Free Reinforcement Learning.

Schwarting, W., Alonso-Mora, J., Rus, D., 2018. Planning and Decision-Making for Autonomous Vehicles. Annu. Rev. Control. Robot. Auton. Syst. 1, 187–210. https://doi.org/10.1146/annurev-control-060117-105157

Sen, B., Smith, J.D., Najm, W.G., 2003. Analysis of lane change crashes. Final Rep. DOT Hs 809 702. https://doi.org/DOT-VNTSC-NHTSA-02-03

Sinha, K.C., Labi, S., 2007. Transportation Decision Making: Principles of Project Evaluation and Programming, Transportation Decision Making: Principles of Project Evaluation and Programming. https://doi.org/10.1002/9780470168073

Suh, J., Chae, H., Yi, K., 2018. Stochastic Model-Predictive Control for Lane Change Decision of Automated Driving Vehicles. IEEE Trans. Veh. Technol. https://doi.org/10.1109/TVT.2018.2804891

Sun, D. (Jian), Elefteriadou, L., 2010. Research and Implementation of Lane-Changing Model Based on Driver Behavior. Transp. Res. Rec. J. Transp. Res. Board 2161, 1–10. https://doi.org/10.3141/2161-01

Sutskever, I., Vinyals, O., Le, Q. V., 2014. Sequence to sequence learning with neural networks, in: Advances in Neural Information Processing Systems.

TRB, 2019. TRB Forum on Preparing for Automated Vehicles and Shared Mobility: Mini-Workshop on the Importance and Role of Connectivity, in: Transportation Research Circular.

TRB, 2018. Socioeconomic Impacts of Automated and Connected Vehicle: Summary of the Sixth EU–U.S. Transportation Research Symposium, in: Transportation Research Board Conference Proceedings.

Treiber, M., Kesting, A., 2013. Traffic Flow Dynamics, Traffic Flow Dynamics. https://doi.org/10.1007/978-3-642-32460-4

USDOT, 2019. Preparing for the Future of Transportation: Automated Vehicles 3.0.

Van Hasselt, H., Guez, A., Silver, D., 2016. Deep reinforcement learning with double Q-Learning, in: 30th AAAI Conference on Artificial Intelligence, AAAI 2016.

Veres, S.M., Molnar, L., Lincoln, N.K., Morice, C.P., 2011. Autonomous vehicle control systems - A review of decision making. Proc. Inst. Mech. Eng. Part I J. Syst. Control Eng. https://doi.org/10.1177/2041304110394727

Wang, P., Chan, C.Y., De La Fortelle, A., 2018. A Reinforcement Learning Based Approach for Automated Lane Change Maneuvers, in: IEEE Intelligent Vehicles Symposium, Proceedings. https://doi.org/10.1109/IVS.2018.8500556

Watkins, C.J.C.H., Dayan, P., 1992. Q-learning. Mach. Learn. https://doi.org/10.1007/bf00992698

World Bank, 2005. A Framework for the Economic Evaluation of Transport Projects, Transport Notes.

Xie, D.F., Fang, Z.Z., Jia, B., He, Z., 2019. A data-driven lane-changing model based on deep learning. Transp. Res. Part C Emerg. Technol. https://doi.org/10.1016/j.trc.2019.07.002

Yang, D., Zheng, S., Wen, C., Jin, P.J., Ran, B., 2018. A dynamic lane-changing trajectory planning model for automated vehicles. Transp. Res. Part C Emerg. Technol.